\documentclass{article}

\PassOptionsToPackage{numbers, compress}{natbib}

\usepackage[preprint]{neurips_2021}

\usepackage[utf8]{inputenc} 
\usepackage[T1]{fontenc}    
\usepackage{hyperref}       
\usepackage{url}            
\usepackage{booktabs}       
\usepackage{amsfonts}       
\usepackage{nicefrac}       
\usepackage{microtype}      
\usepackage[pdftex]{graphicx}
\usepackage[frozencache=true,cachedir=minted-cache]{minted}
\usepackage{xcolor}         
\usepackage[numbers]{natbib}

\title{Converse: A Tree-Based Modular Task-Oriented Dialogue System}

\author{%
  Tian Xie\thanks{Equal contribution.} \\
  Salesforce Research \\
  \texttt{txie@salesforce.com} \\
  \And
  Xinyi Yang\footnotemark[1] \\
  Salesforce Research \\
  \texttt{x.yang@salesforce.com}
  \And
  Angela S. Lin \\
  Salesforce Research \\
  \texttt{angela.lin@salesforce.com} \\
  \And
  Feihong Wu \\
  Salesforce Research \\
  \texttt{feihong.wu@salesforce.com} \\
  \And
  Kazuma Hashimoto\thanks{Work done while at Salesforce Research.} \\
  Google Research \\
  \texttt{freedomigo@gmail.com} \\
  \And
  Jin Qu \\
  Salesforce Research \\
  \texttt{jqu@salesforce.com} \\
  \And
  Young Mo Kang\footnotemark[2] \\
  Microsoft \\
  \texttt{youngmokang@microsoft.com} \\
  \And
  Wenpeng Yin\footnotemark[2] \\
  Temple University \\
  \texttt{wenpeng.yin@temple.edu} \\
  \And
  Huan Wang \\
  Salesforce Research \\
  \texttt{huan.wang@salesforce.com} \\
  \And
  Semih Yavuz \\
  Salesforce Research \\
  \texttt{syavuz@salesforce.com} \\
  \And
  Gang Wu \\
  Salesforce Research \\
  \texttt{gang.wu@salesforce.com} \\
  \And
  Michael Jones \\
  Salesforce Research \\
  \texttt{jones.m@salesforce.com} \\
  \And
  Richard Socher\footnotemark[2] \\
  you.com \\
  \texttt{rsocher@gmail.com} \\
  \And
  Yingbo Zhou \\
  Salesforce Research \\
  \texttt{yingbo.zhou@salesforce.com} \\
  \And
  Wenhao Liu \\
  Salesforce Research \\
  \texttt{wenhao.liu@salesforce.com} \\
  \And
  Caiming Xiong \\
  Salesforce Research \\
  \texttt{cxiong@salesforce.com} \\
}

\begin{document}

\maketitle

\begin{abstract}
Creating a system that can have meaningful conversations with humans to help accomplish tasks is one of the ultimate goals of Artificial Intelligence (AI). It has defined the meaning of AI since the beginning. 
A lot has been accomplished in this area recently, with voice assistant products entering our daily lives and chat bot systems becoming commonplace in customer service. At first glance there seems to be no shortage of options for dialogue systems.
However, the frequently deployed dialogue systems today seem to all struggle with a critical weakness - they are hard to build and harder to maintain. At the core of the struggle is the need to script every single turn of interactions between the bot and the human user.
This makes the dialogue systems more difficult to maintain as the tasks become more complex and more tasks are added to the system.
In this paper, we propose Converse, a flexible tree-based modular task-oriented dialogue system. Converse uses an and-or tree structure to represent tasks and offers powerful multi-task dialogue management. Converse supports task dependency and task switching, which are unique features compared to other open-source dialogue frameworks. At the same time, Converse aims to make the bot building process easy and simple, for both professional and non-professional software developers. The code is available at \href{https://github.com/salesforce/Converse}{https://github.com/salesforce/Converse}.

\end{abstract}

\section{Introduction}

Task-oriented dialogue (TOD) systems help users complete tasks through dialogue \citep{Jurafsky-book}. TOD systems can be valuable for various industries. Examples of TOD tasks include: reserving a table at a restaurant, booking flight tickets, making a customized T-shirt order, or checking vaccination appointment availability.  The typical pipeline for task-oriented dialog systems are divided into several modules and are built on top of a structured task ontology \citep{zhang2020recent}. However, there are several challenges in building task-oriented dialogue systems. 

Firstly, it is challenging to define an efficient task ontology that captures the variation in dialogue. Due to the varied nature of dialogue, there can be several different ways or orders to complete the same task. For example, the user can check the delivery status of an order using the order id, or the username and the delivery address. Ideally, a task ontology should be simple and efficient at capturing all of these task variations. If a bot builder needs to specify all possible trajectories for completing a task, the building process soon becomes intractable due to the combinatorial nature of the task. 

Secondly, the adaptation to new tasks and building a new bot can be costly. For instance, most dialogue state tracking models are domain-specific, and cannot be directly generalized to new domains. For most pipeline TOD systems, introducing a new task or building a new bot requires resources (annotated data, training time, and computational resources) to retrain the natural language understanding module to detect the desired new intent and entities. These resource requirements hamper the bot building process, reducing adoption and impeding bot builders when they try to add new use-cases.

Thirdly, handling conversations with interactions between tasks is difficult. When building a bot, it is standard to treat each task as a standalone configuration. But in real conversations with users, there will be interactions between tasks, for example task switching. Task switching is a situation where a user first starts a task X; then in the middle of task X, the user starts a new task Y; after finishing task Y, the user should be able to pick up where they left off in task X and continue the conversation. Another example is the dependency between tasks. If checking the order status and changing the delivery address both depend on user authentication, then after checking the order status (and also authenticating the user), the dialogue system should not authenticate the user again when changing the delivery address. Solving these challenges is essential for creating a system that works well for users who interact with bots and bot builders who build bots.

To address the challenges above, we propose Converse, a low-code, tree-based, modular, task-oriented dialogue system. Converse defines a task as an and-or tree structure \cite{xiong-etal-2016-and-or} that is called Task Tree. A Task Tree is composed of the key pieces of information (entities and sub-tasks) needed for the task and their relationships. This structure can represent various kinds of tasks. The system traverses the tree to easily reference, maintain, and update the task-related dialogue states. The bot builders\footnote[1]{To avoid confusion on the persona involved in this paper, we use the following definitions. A bot builder (chatbot builder or bot admin) is the person who creates chatbots using Converse. A user is a person who interacts with the runtime bot system. A user session is created every time when a user starts a new conversation with the chatbot.} do not need to define the next step after each dialogue state one by one. Instead, they only need to define the Task Tree structure. The tree traversal automatically traverses the Task Tree and finds the next entity (slot) and next task. Furthermore, the Task Tree also takes care of interactions between tasks in conversations, such as switching tasks and completing dependent sub-tasks. The Task Tree is versatile and simplifies the bot building process.

Similarly, the other components of Converse can be utilized with ease. The bot builder can build a new bot by writing a few configuration files and a bit of code. To configure a bot, the bot builder only needs to define Task Trees for their tasks that include: the information to collect from users and actions to take after information is collected. To use the natural language understanding module, annotated data or training models on in-domain data are not required. We introduce a natural language inference based few-shot intent detection model, which enables the bot builder to add new intents by listing a few samples. We also provide common predefined entity types that the user can use out-of-the-box without collecting more training data and re-training the model. If bot builders want to further customize the natural language understanding module, Converse's modular design enables them to smoothly switch out our models with their own. These design choices save bot builders effort and time.


In this paper, we will mainly focus on how the tasks are represented in Converse, i.e., the Task Tree and the Dialogue Management.

\section{Converse's Abilities and Limitations}

\subsection{What Converse can do}

Before we talk about the technical details of Converse, we want to first show what Converse can do and what Converse cannot do. We also provide an example configuration of a Converse bot in \ref{example-bot}.

\subsubsection{Define and reuse sub-tasks}

Let's begin with an online shopping helper bot which can help the user check their order status and update their order. For security, the bot will verify the user's identity first before starting these two tasks. 

\emph{\textbf{User}: Hi, I would like to check my order status.}

\emph{\textbf{Bot}: Oh sure, I'd be happy to help you check your order status. First, I need to pull up your account. Could you please tell me your email address?}

Checking the order status is the \textbf{main task}, which has a \textbf{sub-task} to verify the user's identity. We also used this sub-task in the task for updating the order. Bot builders only need to define a sub-task once. Then the sub-task can be used in other tasks. Reusing sub-tasks reduces the amount of configurations that the bot builder needs to write. In addition, reusing sub-tasks allows Converse to skip the sub-task if it has already been completed which improves the user experience, for example, verifying the user's identity only once in a session.

\subsubsection{Use and-or relations in the Task Tree to handle branching conversations}

The following conversation is a continuation of the example above. In this conversation, the bot adjusts its response based on whether the user provides the correct email address.

\emph{\textbf{Bot}: Oh sure, I'd be happy to help you check your order status. First, I need to pull up your account. Could you please tell me your email address?}

\begin{tabular}{ p{7cm}|p{7cm} } 
 Trajectory 1 & Trajectory 2 \\ 
  \hline
\emph{\textbf{User}: [the correct email address]} &
\emph{\textbf{User}: I don't remember it.} \\

\emph{\textbf{Bot}: I have verified your identity. Please provide your order id for your order status.} &

\emph{\textbf{Bot}: I am sorry, but I could not recognize your email address. Could your please tell me your zip code?} \\

 & \emph{\textbf{User}: [the correct zip code]} \\

& \emph{\textbf{Bot}: I have verified your identity. Please provide your order id for your order status.} \\ 
\end{tabular}

Note that the bot builders only need to create the and-or relations, instead of specifying every possible linear conversational path. The Task Tree in Converse is able to handle more nested and complex structures than in the above example, as long as they can be represent in and-or relationships.

\subsubsection{Switch between tasks in the middle of a task}

Here's an example of a flight booking bot where we'd like to show Converse's multi-task conversation handling.

\emph{\textbf{User}: I'd like to book a round-trip flight.}

\emph{\textbf{Bot}: I'd be happy to help you book a flight. Where will you depart from?}

\emph{\textbf{User}: Oh wait, could you please help me check tomorrow's weather first?}

\emph{\textbf{Bot}: I'd be happy to help you check the weather. What is the zip code of your area?}

The user started a flight booking task, but after the bot asked about the departure location, the user changed the mind and started a new weather query task. The bot detected the new intent, \textbf{switched to the new task} weather query, and asked for the zip code.

\emph{\textbf{User}: [a zip code].}

\emph{\textbf{Bot}: The weather in [city name of the zip code] is [some weather]. That's all I have about the weather. Where will you depart from?}

The bot successfully helped the user to query the weather. Then it picked up what it left from the previous conversation, \textbf{switched back} to flight booking task, and asked the user about the origin location again.

\subsubsection{Configure responses to frequently asked questions}

Converse can not only switch between tasks, but also answer frequently asked questions (FAQs). The following conversation continues from the previous conversation:

\emph{\textbf{User}: San Francisco.}

\emph{\textbf{Bot}: Got it. Where is your destination?}

\emph{\textbf{User}: Do I have free checked bags?}

\emph{\textbf{Bot}: All frequent flyer program members will have one free checked bag. Where is your destination?}

Converse defines \textbf{FAQs} as one-turn dialogues that can happen anywhere in a conversation. FAQs can be easily configured in the task configuration by listing the question and answer pairs, such as \emph{"Do I have free checked bags? - All frequent flyer program members will have one free checked bags."}. Here the bot detected the question from the conversation and fetched the answer from FAQ configuration, then continued the flight booking task.

\subsection{What Converse cannot do}

\subsubsection{Multilingual support}
Currently, Converse only supports English. If Bot builders want to build chatbots in other languages, they need to use their own NLU models and response templates.

\subsubsection{Canceling an ongoing task}
Currently, Converse doesn't support canceling an ongoing task. Here is an example conversation:

\emph{\textbf{Bot}: Hi there, I am the digital assistant for Northern Trail Information Center. What can I do for you?}

\emph{\textbf{User}: I want to check order status}

\emph{\textbf{Bot}: Oh sure, I'd be happy to help you check your order status. First, I need to pull up your account. What is your email address?}

\emph{\textbf{User}: I don't want to check order status anymore}

\emph{\textbf{Bot}: I am sorry, but I could not recognize your email address. What is your zip code?}

In this example, the user changes the mind, and doesn't want to continue the check order status task anymore. However, the bot is still trying to extract entity (slot) information from the user. 

\subsubsection{Discriminate between entities that have the same entity type}
Here's an example of a flight booking conversation that multiple entities with the same entity type mentioned in one user's utterance.

\emph{\textbf{User}: I'd like to book a round trip flight from San Francisco to Los Angeles for 2 people.}

\emph{\textbf{Bot}: I'd be happy to help you book a flight, I got multiple possible answers for origin: San Francisco and Los Angeles, which one did you mean? Could you walk me through the details?}

In above conversation the user already specified the origin and the destination. Unfortunately, Converse detects entity based on type instead of name; it treats San Francisco and Los Angeles equally as location entities and fit them both into the first required location entity \emph{origin}. 

\subsubsection{Commonsense Reasoning}
Converse can handle only a limited amount of commonsense reasoning. In some cases, this may cause the bot to incorrectly process user input. Here’s an example:

\emph{\textbf{Bot}: And here are your returning flights: \\
                    Oceanic 443, Depart at 4:16am, 800 USD \\
                    Ajira 232, Depart at 15:42pm, 1500 USD \\
                    Qantas 424, Depart at 23:08pm, 2300 USD} 
                    
\emph{\textbf{User}:  the late night one} 

\emph{\textbf{Bot}: Alright, your returning flight is Oceanic 443.} 

In this case, the bot is not able to relate “late night” to “23:08pm”. However, the bot can understand “the last one” or “Qantas 424” based on the rules we have implemented. 

\subsubsection{Mixed initiatives}
In Converse, the bot always leads the conversations. Sometimes, users may not follow the bot, and provide something not expected by the bot. Here’s an example:

\emph{\textbf{Bot}: Hi there, I am the digital assistant for Northern Trail Information Center. What can I do for you?}

\emph{\textbf{User}: I want to check order status}

\emph{\textbf{Bot}: Oh sure, I'd be happy to help you check your order status. First, I need to pull up your account. What is your email address?}

\emph{\textbf{User}: Can I use my name to verify my identity?}

\emph{\textbf{Bot}: Oh sure, I'd be happy to help you check your order status. First, I need to pull up your account. What is your email address?}

Converse doesn't allow users to lead the conversations, like other existing task-oriented dialogue system frameworks, hence the conversations may not be as natural as human conversations when the users don’t follow the bot.

\subsubsection{Talk to users about open-ended topics}
Converse is not an open-domain dialogue system. The conversations are restricted by the defined tasks. Bot builders can define FAQs to make Converse able to handle chitchat. Since Converse doesn't have a model-based Natural Language Generation (NLG) module, Converse’s chitchat ability is limited by the FAQs defined by bot builders. 

\section{Related Work}
\label{gen_inst}

In recent years, there has been an increasing amount of literature on studying task-oriented dialogue systems. There are two main directions for this research: (1) frameworks for building task-oriented bots \citep{bocklisch2017rasa, plato-dialogue-system, botpress, dialogflow, bot-framework-composer, watson, lex, power-virtual-agents}, and (2) systems for evaluating new dialogue models \citep{ultes-etal-2017-pydial,miller-etal-2017-parlai}. Converse aligns with the first research direction with a focus on creating a task-oriented dialogue system that does not require much code to set up.

In existing conversational frameworks, bot builders need to specify every user intent and bot action in a conversation, giving bot builders fine-grained control over the bot’s behavior at the cost of simplicity. In Rasa \citep{bocklisch2017rasa}, bot builders create a list of user intents and bot actions for every linear conversational path that the bot handles. However, natural conversations branch when the user changes the topic or does not respond to the bot’s questions, which does not match Rasa's linear conversation configuration design. In contrast, bot builders using Botpress \citep{botpress}, Google Dialogflow \citep{dialogflow}, Microsoft Bot Framework Composer \citep{bot-framework-composer}, and IBM Watson Assistant \citep{watson} connect blocks that represent user intents and bot actions in a graphical interface to configure the conversational flow of a task using a graph. These conversational graph structures are more flexible to handle branches in the conversation, however, bot builders still need to specify the bot's behavior at every conversational step.  In Converse, on the other hand, bot builders specify the entities for each task and the Task Tree structure, and the built-in dialogue policy handles generating responses based on the dialogue states, switching tasks, and steering the conversation back on track when the user does not answer the bot’s questions, simplifying the conversation configuration compared to existing conversational frameworks.

The task structure used in Converse is similar to a feature in Rasa, Amazon Lex, and Google Dialogflow that collects multiple entities from users that are required to complete a task, such as the date and time of an appointment. In contrast to these other dialogue systems where bot builders can only configure entities as required or optional, bot builders can use the task structure in Converse to configure more complex logic to collect entities, such as asking the user to provide two out of three types of user authentication (e.g. their zip code, a two-factor authentication code, and government ID number).

\section{System overview}
\label{headings}

Modern task-oriented dialogue systems are usually designed to be modular or a pipeline. They usually comprise modules such as Natural Language Understanding (NLU), Dialogue State Tracking (DST), Dialogue Management (DM), and Natural Language Generation (NLG). To solve real-world problems, a Knowledge Base (KB)/ Backend action is often connected to the task-oriented dialogue system. The Knowledge Base can be a database, a knowledge graph, or anything else that contains necessary information to fulfill user’s inquiry. In addition to the aforementioned modules, fully functional spoken dialogue systems also incorporate Automatic Speech Recognition (ASR) and Text to Speech (TTS) modules.

\begin{figure}[h]
    \centering
    \includegraphics[width=1\linewidth]{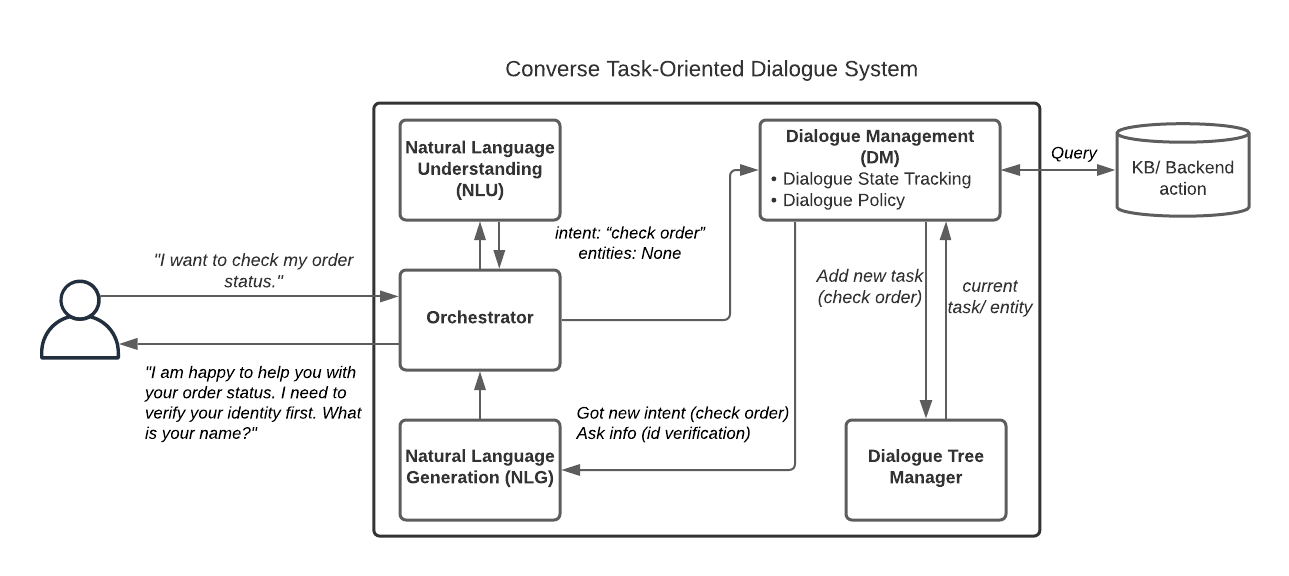}
    \caption{The architecture of Converse task-oriented dialog system}
    \label{fig:tod_sys}
\end{figure}

Converse follows the typical design of modern task-oriented dialogue systems, see Figure \ref{fig:tod_sys}. The Orchestrator module works as the receptionist of Converse. It handles communications between the user and the system, and also coordinates with other modules. The core of Converse is Dialogue Management (DM). The DM is connected with a unique component in Converse called the Dialogue Tree Manager. The Dialogue Tree Manager manages the Task Tree, which is an and-or tree-based data structure that represents tasks in Converse.

More details about the NLU module and the NLG module can be found in the appendix and the documentation in the GitHub repo.

\section{And-Or Task Tree}
\label{others}

\subsection{A new way to represent tasks}

In task-oriented dialogue systems, tasks (intents) are the specific things that we want the system to help users achieve, like checking the weather or booking movie tickets.  A task consists of a collection of slots (entities) and their relationships. The simplest case is that a task has several necessary entities. To complete the task, the bot needs to ask the user for the entity value one by one. In more complex cases, the bot may only need to ask the user for some entity value when a condition is met. In other words, to complete a task in different situations, different combination of entities can be used.

The representation of tasks in a task-oriented dialogue system determines how the system handles the tasks. In other words, the representation of tasks is the organization of the slots (entities) for this task.

There are different ways to represent tasks in a task-oriented dialogue system. In Converse, we chose an and-or tree structure to represent tasks, instead of creating a story-based conversation turn by turn in a linear structure. We defined a Task Tree structure based on and-or tree. 

Why do we use and-or tree structure? We want to formulate every task as a problem to solve, and make it possible to be solved in a generic way. The problem can be considered as a combination of a set of sub-problems. The and-or tree structure represents the search space for solving the problem. Different search strategies for searching the space are possible. In Converse, the Dialogue Tree Manager traverses the Task Tree based on the dialogue states that are managed by the dialogue management module.

\subsection{How to use and-or tree to represent tasks}

With the and-or tree structure, we can convert almost every task into a combination of and/or relations. For example, if a task verifies the user’s identity, and we can verify either by name and email address, or by birthday and social security number (SSN), the task can be represented as (Verify \emph{name} and Verify \emph{email address}) or (Verify \emph{birthday} and Verify \emph{SSN}).

The and-or tree structure is simple yet powerful for representing tasks. We don’t need to create different stories for different tasks. We only need to reduce the task into several basic parts, and connect them with and/or relations. In this way, configuring a new task can be much easier for bot builders.

We defined 3 type of tree nodes for the Task Tree: And node, Or node, and Leaf node. Each node has a success status label. If a tree node’s success status label is \emph{True}, it means this node is successfully finished based on the information extracted from the previous conversation, and it won't be visited again later in the conversation. For each type of node, we define a node success condition to decide the value of success status label, which by default is \emph{False}. 

The child nodes of an And node or Or node can be And nodes, Or nodes and Leaf nodes. Leaf nodes are entity nodes, which store a group of entities and the extracted entity values. 

\begin{table}[h]
  \caption{Node types and success conditions}
  \label{table:task_tree_node}
  \centering
  \begin{tabular}{ll}
    \toprule
    Node Type        & Success Condition \\
    \midrule
    And node  & all the child nodes’ success labels are \emph{True} \\
    Or node  &  at least one child node’s success label is \emph{True}  \\
    Leaf node  & all or a specified number of entities have correct values \\
    \bottomrule
  \end{tabular}
\end{table}

At the beginning of each conversation, the Task Tree is a root node. The root node is an Or node by default. The system don’t check the root node’s success status. It only checks the task nodes. The task nodes (the tree nodes represent tasks) are And or Or nodes. The task nodes will be added to the Task Tree as the conversation goes.

\subsection{The benefits of using an and-or tree to represent tasks}

\begin{figure}[h]
    \centering
    \includegraphics[width=1\linewidth]{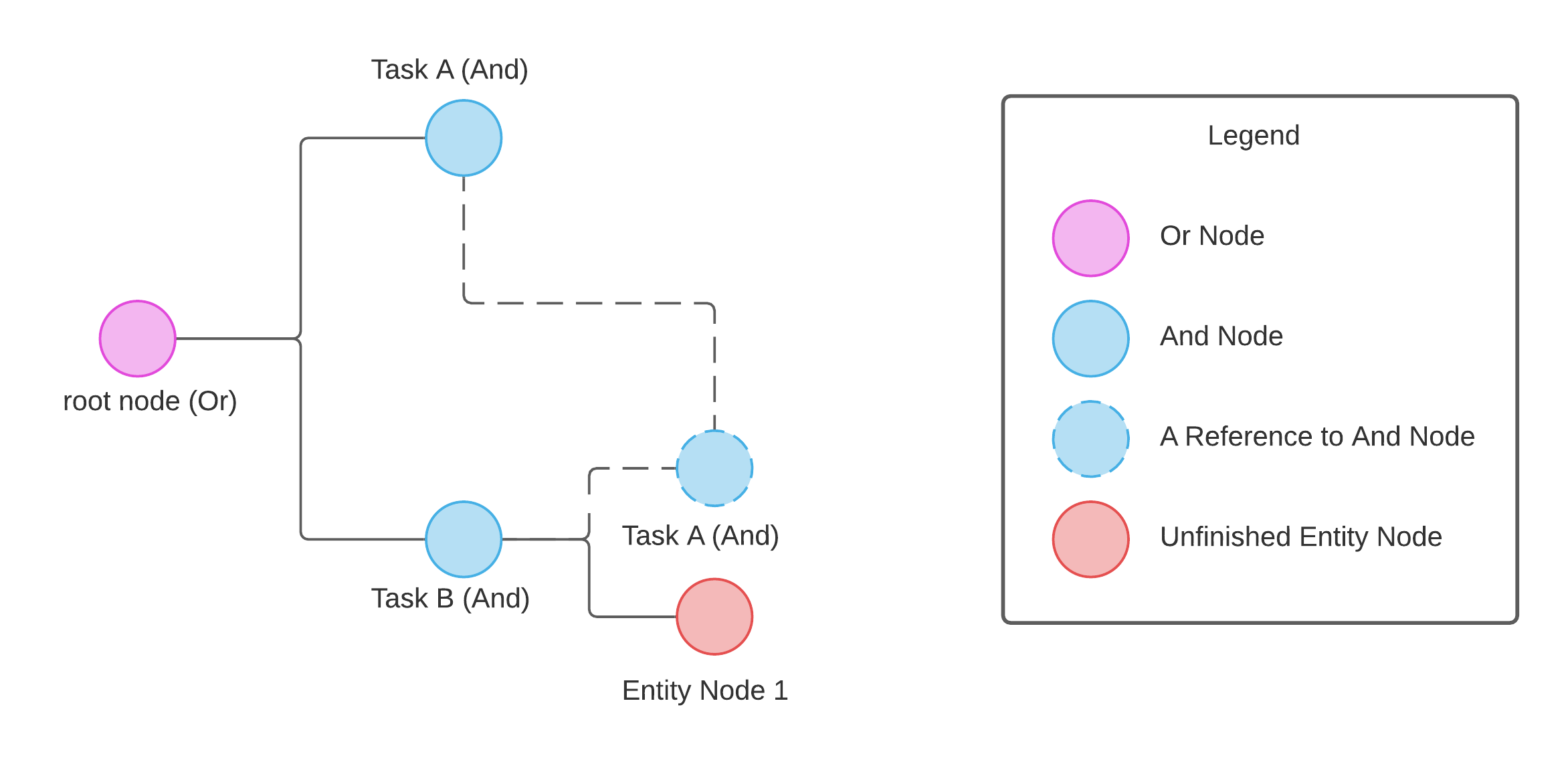}
    \caption{The Task Tree visualization of a task with sub-task}
    \label{fig:task_tree_sub_task}
\end{figure}

With the tree nodes we defined, the bot builders can configure complex and/or relations, which means they can build complex task structures easily. A task can depend on other tasks, i.e., it can have prerequisite tasks (sub-tasks). A sub-task can be shared by different tasks. For example, bot builders may configure several tasks and they both need identity verification. Then identity verification can be a shared sub-task. If task A is task B’s sub-task, Task node A should be Task node B's child node intuitively. Note that in actual implementation, in order to only have one copy for each task and check task success status easily, all the task nodes are sibling nodes with the same parent node, which is the root node. If task A is task B’s sub-task, task node B won't actually have task node A as its child node, but a reference to task node A. In this way, task node A is still the root node's child node, and task node B's structure is well-maintained, see Figure \ref{fig:task_tree_sub_task}.

The Task Tree supports task switching. Users can tell the bot they want to do task B when the bot is currently handling task A. The bot will pause task A and switch to task B. After task B is finished, the bot will go back to task A, and continue working on task A from where it stopped.

The Task Tree supports task repetition if the task is reusable. When a task is repeated, the task node and all its child nodes will be reset. However, if the task has sub-tasks, the sub-tasks will not be reset. For example, if a task is \emph{making online order}, and it has a sub-task, \emph{verify identity}. The \emph{making online order} can be repeated to make another order, but the bot only needs to verify the user’s identity when making the first order. Of course, if sub-tasks are also independent tasks, they can be reset if users say they want to do it again.

\subsection{How Task Tree is managed in Converse}

The Task Tree Manager in Converse handles operations on the Task Tree, including adding a new task, tree traversal, and quitting a task. The Task Tree Manager maintains the current task, current entity name, and the current node. The current node is always a Leaf node. Adding a new task will construct the task structure and add it under the root node. When the current node is finished (either the success label is True or all the entities have been tried), tree traversal can traverse to the next unfinished Leaf node. When the current node is unfinished, tree traversal will stay at the current node. The Task Tree manager quits a task when the number of turns about the task exceeds the limit. The Task Tree Manager interacts with Dialogue Management module, as shown in Figure \ref{fig:tod_sys} previously.

\begin{figure}[h]
    \centering
    \includegraphics[width=1\linewidth]{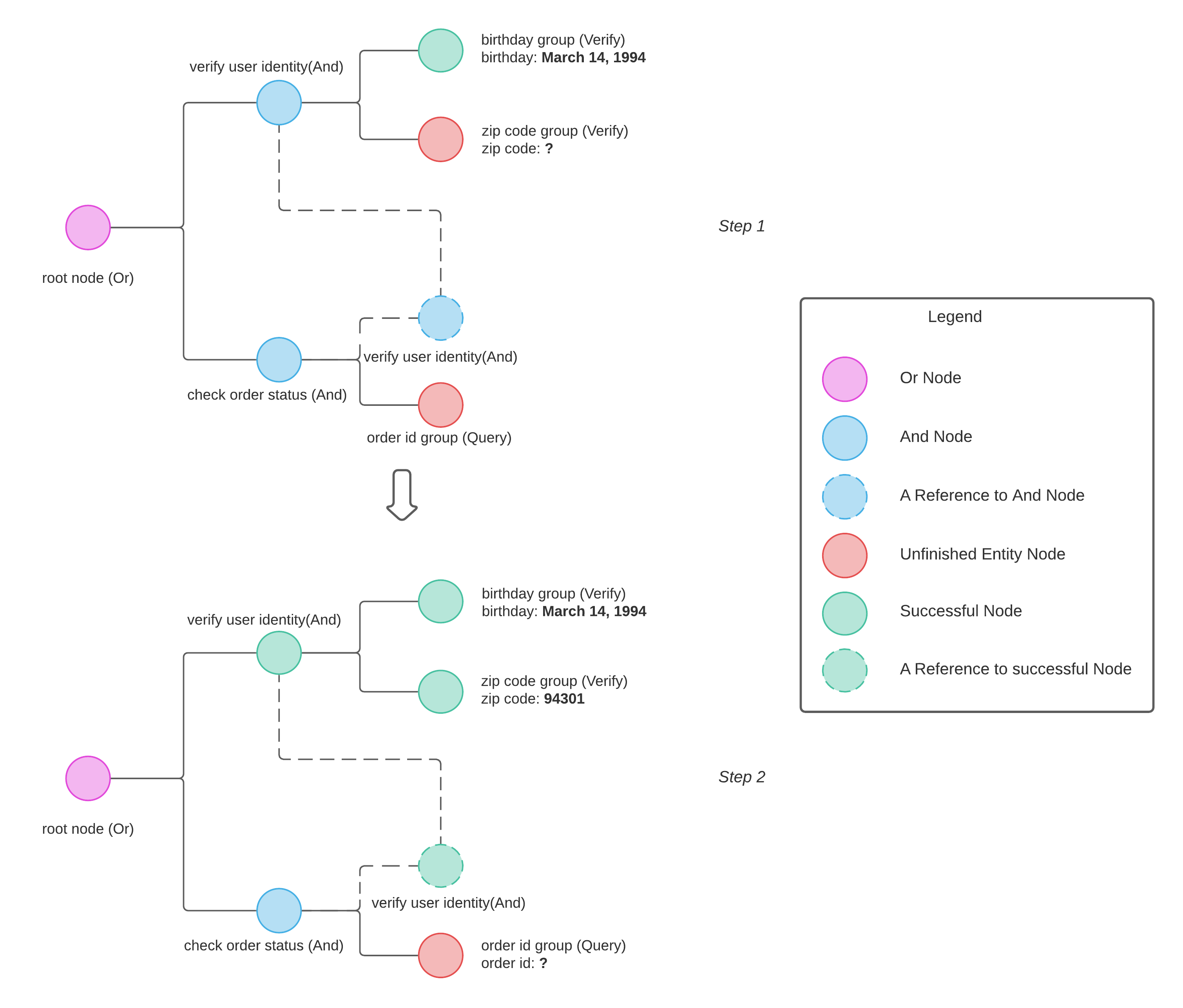}
    \caption{An example of Task Tree operations - check order status}
    \label{fig:task_tree_check_order}
\end{figure}

Figure \ref{fig:task_tree_check_order} shows an example on how the Task Tree Manager works in detail. There are two tasks in the Task Tree,  \emph{check order status} and \emph{verify user identity}. The bot will work on the sub-task (\emph{verify user identity}) first. In \emph{Step 1}, for task \emph{verify user identity}, the information of slot \emph{birthday} has been verified, and the current slot or entity is \emph{zip code}. When the sub-task is successfully finished, the Task Tree Manager will use tree traversal operation to switch to the main task (check order status). After this operation, the current slot or entity will be \emph{order id}, see \emph{Step 2}.

\section{Dialogue Management}



\subsection{Dialogue State Tracking}

The dialogue state is the representation of the system's beliefs about the task(s) at any time during the dialogue given the dialogue history. Dialogue state tracking (DST) is a key component in Dialogue Management that extracts user-provided key information during the conversation and encodes them as a compact set of dialogue states \citep{WuTradeDST2019}.

In Converse, there are three kinds of dialogue states: single-turn states, multi-turn states, and task-related states.

Converse uses single-turn states to know what's going on in the current turn. The states include whether intent or entity information is extracted in this turn, the polarity (positive, negative, neutral) of the user utterance, etc.

Converse uses multi-turn states to remember what happened in the previous conversation turns to generate relevant dialogue in the current turn. With the multi-turn states, Converse is able to reuse the information the user already provided in the previous turns. For example, if the system tried to confirm something with the user in the previous turn, a yes-or-no answer is expected in the current turn. Another example is if a conversation length exceeds the turn limit, then it might be time to try other solutions like connecting to a human agent. 

Converse uses task-related states to represent the system's belief of the tasks. The Task Tree itself maintains the task-related states. It tells the system the current task and the current entity to focus on. The Task Tree Manager also stores the tasks that were finished and are waiting to be finished.
 
Converse uses all the three kinds of dialogue states to track the flow of dialogue. The dialogue states accurately identify what the user wants and what process the system is in. The dialogue states are managed by two modules in Converse - the Orchestrator and the Dialogue State Manager. In each turn of a conversation, the Orchestrator coordinates with other modules to collect and update the single-turn and multi-turn states. If any operations on the tree are triggered by these updated dialogue states (decided by the Dialogue Policy and will be explained in detail later in the paper), the Dialogue State Manager will update the Task Tree via Task Tree Manager.

\subsection{Dialogue Policy}

The Dialogue Policy in Converse follows a human-designed policy graph (which can be customized) to decide the next dialogue action based on the dialogue states, and generate the response through the Response Generation module. The policy graph is a decision tree, which we call the Policy Decision Tree.

When the Dialogue Policy receives the updated dialogue states, it follows the Policy Decision Tree to find the next dialogue action. The leaf nodes in the Policy Decision Tree are the names of dialogue actions. All the other nodes in the Policy Decision Tree are decision rules or conditions. During the execution of the Dialogue Policy, Converse traverses the Policy Decision Tree through a depth-first search (DFS). When a node is a condition, if the condition is True, then its children can be visited. The DFS stops when a leaf node is reached. Then the corresponding dialogue action will be executed.

The Dialogue Policy is shared between all the tasks in Converse. We assume that all the tasks can be done by a set of general dialogue actions. The bot builders don't need to write dialogue actions by themselves for the tasks they defined. The dialogue actions can be simple or complex, depending on simple or complex cases we want them to deal with. An example of a simple dialogue action is a greetings action, which generates a greeting response like "Hi, how are you?" when conditions are met. An example of a complex dialogue action is a \emph{new task with info} action, which handles the conversation when both intent and entity information are extracted from user input. For a user's input like "I would like to check the weather for zip code 94105.", the \emph{new task with info} action will add the new \emph{check the weather} task to the Task Tree, update the \emph{zip code} entity node on the Task Tree, make the API call to query the weather and finally generate a response "I would be happy to help you query the weather, the weather in 94105, San Francisco is sunny. Is there anything else I can help you with?".

With the Dialogue Policy, bot builders do not need to create different conversation flows for different tasks. Sometimes, bot builders may find their use cases require capabilities beyond what the current Dialogue Policy can do. They can add their dialogue action code, then modify the conditions and corresponding dialogue action names in the Policy Decision Tree configuration file.

\begin{figure}[h]
    \centering
    \includegraphics[width=1\linewidth]{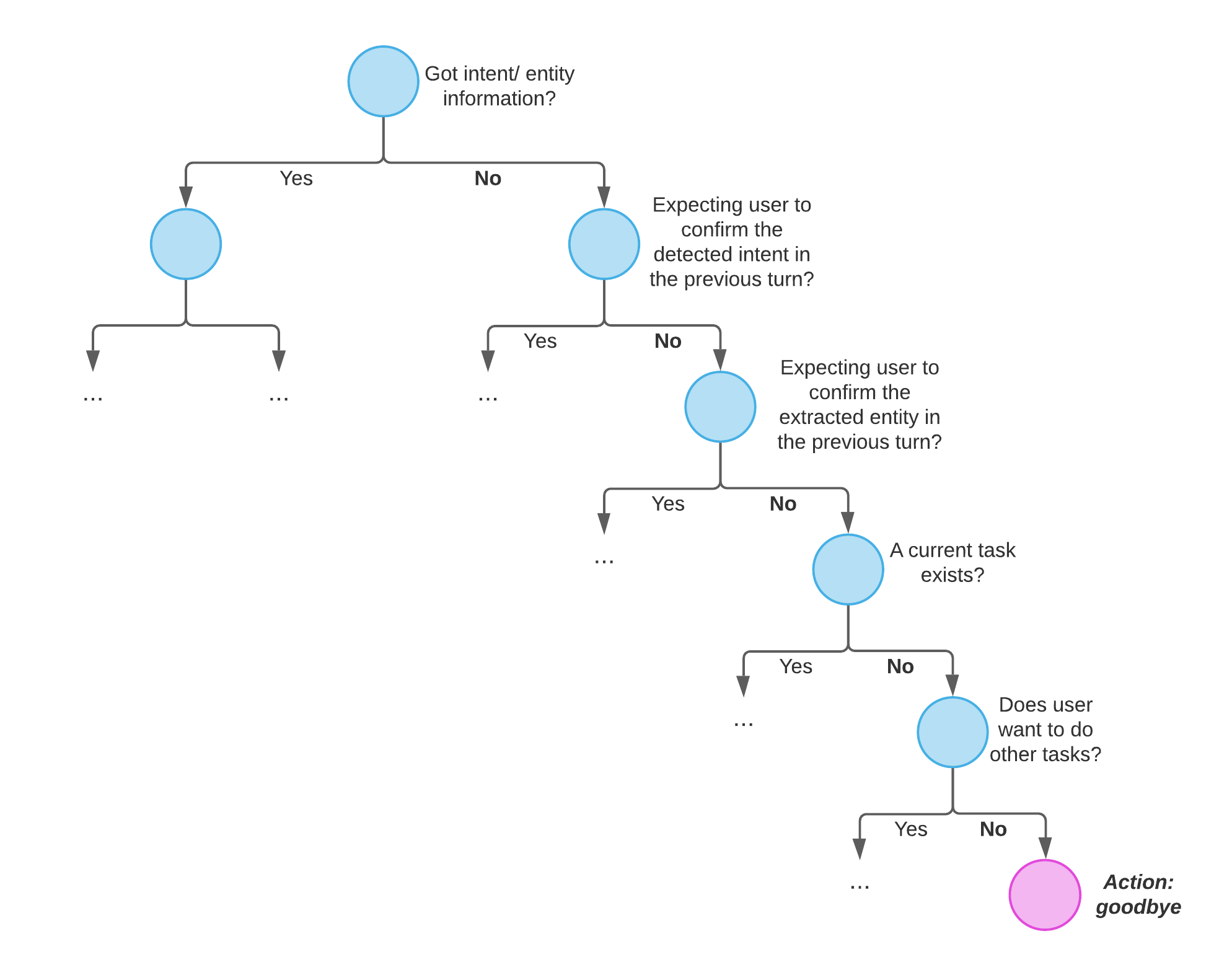}
    \caption{An example of the execution of the Policy Decision Tree}
    \label{fig:policy}
\end{figure}

Here, we show an example of the execution of the Policy Decision Tree, see Figure \ref{fig:policy}.

\section{Ethical Considerations}
Chatbot systems can be capable of producing unethical conversations. One example is Microsoft’s Tay which learned and replicated offensive language within hours of interacting with users on Twitter \citep{wolf2017we}. However, because Converse does not contain a model-based NLG model, this capability is mitigated in our system. A bot builder could still provide biased, racist, or otherwise offensive Task Tree, training samples, or FAQs resulting in inappropriate responses, but Converse does not have the ability to learn offensive or racist language from the end-users. Likewise, Converse does not have the ability to learn and distribute PII, keeping a user’s information private. As with other chatbot systems, there is a potential for dialogue failure. To minimizes the risk of user frustration and attrition, bot builders can provide contingency plans for users such as links to a help page, or transference to a human operator.

\section{Conclusion}

We introduce Converse, a low-code, tree-based, modular, task-oriented dialogue system. Converse has a powerful and-or-tree task structure to handle various tasks, can be adapted to new tasks easily, and handles inter-task level conversation elegantly. In future work, we will integrate more advanced conversational research into Converse, for example, model-based dialogue policy, open-domain dialogue, dialogue act classification, etc. We will continue to develop Converse as an easy-to-build, easy-to-use, easy-to-deploy, task-oriented dialogue system.

\begin{ack}
The initial idea and overall design for Converse was from Caiming Xiong and Richard Socher. Tian Xie, Kazuma Hashimoto, Jin Qu and Yingbo Zhou built the first PoC, which was demo’ed at \href{https://www.youtube.com/watch?v=3fAuOV6ahqc}{Dreamforce 2019}. Tian Xie, Xinyi Yang, Wenhao Liu, and Caiming Xiong redesigned the architecture of Converse dialogue management core to make the system more flexible, modular, and extensible. Tian Xie, Xinyi Yang, Angela S. Lin, Feihong Wu, and Young Mo Kang further extended Converse by adding additional features and improving ease of setup and deployment. Wenpeng Yin, Huan Wang and Semih Yavuz made significant contributions to the NLU models. We thank Gang Wu for his help in the front-end development. We also thank Michael Jones for his help in product management. Wenhao Liu and Yingbo Zhou led the initiative as project and team managers.  Last but not least, we received input and support from our Einstein product and engineering team members throughout the process. 
\end{ack}

\bibliographystyle{plainnat}
\bibliography{cite}

\appendix

\section{Appendix}

\subsection{Natural Language Understanding (NLU)}
\textbf{Intent detection}

We use a simple and effective approach \citep{zhang-etal-2020-discriminative}, discriminative nearest neighbor classification (i.e., k-nearest neighbors (kNN) classification with $k = 1$), which is based-on natural language inference (NLI). When bot builders configure tasks for Converse, they can add 3 $\sim $ 5 sample user utterances for each task. During bot runtime, when there's a user utterance, the system will compute the entailment scores between the user utterance and all the sample utterances. The sample utterance with the highest entailment score will be selected, and the corresponding task will be the intent detection result.

We also set a threshold for the inference score, usually 0.6 $\sim $ 0.8 to tackle the OOS intent problem.

\textbf{Named-entity recognition (NER)}

We trained an NER model based on TinyBERT$_{6}$\citep{jiao-etal-2020-tinybert}, and we use it as an important entity extraction method for slot filling.

\textbf{Negation detection}

Usually, it is easier to know what the user wants to do in intent detection, while it can be very hard to know what exactly the user doesn't want to do in a dialogue system. It is fairly common to have a system that treats “X does Y” and “X does not Y” as having the very nearly same meaning.\citep{white-2012-uwashington}

Here, we introduce negation detection to help us solve the problem. We defined an algorithm to do intent resolution based-on the results of intent detection model and negation detection model. A coreference resolution model is optional for intent resolution.

\subsection{Dialogue Context}

A conversation from the beginning to the end usually consists of many turns. Each turn is not isolated from the others, which imposes one of the hard challenges on AI bots as they often fail to generate dialogue consistent with past turns. In addition, Converse is designed as a distributed bot system that can serve multiple users at the same time. It needs to maintain multiple conversations and switch from one to the other, therefore the context of every conversation a.k.a "Dialogue Context" should be memorized and managed. These functionalities are fulfilled by the Dialogue Context Manager.

The Dialogue Context Manager can be configured to run in stand-alone mode or distributed mode, with context stored in memory or Redis, respectively.
Also, we implemented the dialogue context manager using the factory design pattern and proxy interface to make it flexible to extend beyond our two implementations.

\subsection{Entity Module}

An entity is a piece of information that the user provides to the system in order to carry out certain tasks. In Converse, we implemented entities as objects instead of using strings because structured objects are easier to manipulate than unstructured text. We provide a dedicated Entity Manager to manage entity operations. We also provide multiple entity extraction methods. 

Unlike most existing dialogue system frameworks, where entities have a life span of a single turn,  entities in Converse are long-lived throughout the user session. We provide an Entity History Manager to manage all the entities extracted throughout the user session and allows the system to insert, remove, and retrieve an entity. This behavior mimics a human agent who can remember entities provided by the user in earlier conversations and does not ask for the same entity already provided in the past.

\subsection{Natural Language Generation (NLG)}

We use a simple template-based NLG module to generate natural language responses. The NLG module connects to the Dialogue Policy. There are two kinds of responses: general responses for all tasks and task-specific responses. Bot builders can configure general responses in a response template file. Task-specific responses can be configured in the task configuration file.

\subsection{Entity backend}

In order to solve real-world problems, we provide entity backend to process the extracted entity values and return messages to the Dialogue Management after processing. For example, if the entity backend is a weather checking API, the input entity value can be a zip code and the return message can be a string contains the weather checking results for this location.

The entity backend can be Python functions or RESTful APIs. In the task configuration file, you can specify the entity backend function/ URL for each entity.

\subsection{Example Converse chatbot} \label{example-bot}

Here, we provide an example Converse bot, which is for making health appointment. To learn more about how to build a chatbot using Converse, please check our \href{https://github.com/salesforce/Converse}{GitHub repo}.

Task configuration YAML file:
\begin{minted}{yaml}
Bot:
  text_bot: true
  bot_name: Nurse Nancy
Task:
  positive:
    description: polarity
    samples:
      - "Yes"
      - "Sure"
      - "correct"
      - "No problem"
      - "that's right"
      - "yes please"
      - "affirmative"
      - "roger that"

  negative:
    description: polarity
    samples:
      - "No"
      - "Sorry"
      - "No, I don't think so."
      - "I dont know"
      - "It's not right"
      - "Not exactly"
      - "Nothing to do"
      - "I forgot my"
      - "I forgot it"
      - "I don't want to tell you"

  get_health_insurance_info:
    description: get health insurance info
    entities:
      ssn:
        function: verify_ssn
        confirm: no
        prompt:
          - May I have the last four digits of your social security number?
        response:
      birthday:
        function: verify_ssn
        confirm: no
        prompt:
          - And your birthday?
        response:
    entity_groups:
      ssn_group:
        - ssn
      birthday_group:
        - birthday
    success:
      AND:
        - VERIFY:
          - ssn_group
        - VERIFY:
          - birthday_group
    finish_response:
      success:
        - I have found your health insurance record.
      failure:
        - Sorry, I am not able to find your health insurance record.
    repeat: false
    max_turns: 10

  health_appointment:
    description: make an appointment at Nurse Nancy
    samples:
      - I want to see a doctor
      - I don't feel so good
      - I would like to make an appointment
      - I want to make an appointment
      - I got a fever
      - make an appointment at Nurse Nancy
    entities:
      appt_date:
        function: collect_info
        confirm: no
        prompt:
          - What date do you prefer for the appointment?
        response:
      appt_time:
        function: collect_info
        confirm: no
        prompt:
          - At what time?
        response:
      department:
        function: collect_info
        confirm: no
        prompt:
          - Which department do you want to make the appointment with?
        response:
      doctor_name:
        function: collect_info
        confirm:
        prompt:
          - May I have your preferred doctor name?
        response:
      covid_protocol:
        function: covid_protocol
        confirm: no
        prompt:
        response:
          - <info>
      have_health_insurance:
        function: check_condition
        confirm: no
        prompt:
          - Do you have health insurance?
        response:
          - <info>
      name:
        function: collect_info
        confirm: no
        prompt:
          - Since you don't have health insurance, let me create a profile for you. 
            What's your name?
        response:
      birthday:
        function: collect_info
        confirm: no
        prompt:
          - What is your birthday?
        response:
          - I have created a profile for you.
    entity_groups:
      date_time_group:
        - appt_date
        - appt_time
      department_doctor_group:
        - department
        - doctor_name
      covid_protocol_group:
        - covid_protocol
      have_health_insurance_group:
        - have_health_insurance
      name_birthday_group:
        - name
        - birthday
    success:
      AND:
        - INSERT:
          - date_time_group
        - INSERT:
          - department_doctor_group
        - OR:
          - AND:
            - API:
              - have_health_insurance_group
            - TASK:
              - get_health_insurance_info
          - INSERT:
            - name_birthday_group
        - INFORM:
          - covid_protocol_group
    finish_response:
      success:
        - I have booked an appointment for you.
      failure:
        - Sorry, I can't help you book an appointment.
    task_finish_function: create_appointment
    repeat: true
    repeat_response:
      - Would you like to make another appointment?
    max_turns: 10
\end{minted}

Entity configuration YAML file:
\begin{minted}{yaml}
Entity:
  ssn:
    type: CARDINAL
    methods:
      ner:
  birthday:
    type: DATE
    methods:
      ner:
  appt_date:
    type: DATE
    methods:
      ner:
  appt_time:
    type: TIME
    methods:
      ner:
  department:
    type: PICKLIST
    methods:
      fuzzy_matching:
        - ICU
        - Elderly services
        - Diagnostic Imaging
        - General Surgery
        - Neurology
        - Microbiology
        - Nutrition and Dietetics
    suggest_value: yes
  doctor_name:
    type: PERSON
    methods:
      ner:
  name:
    type: PERSON
    methods:
      ner:
  covid_protocol:
    methods:
  have_health_insurance:
    type: USER_UTT
    methods:
      user_utterance:
\end{minted}

\begin{figure}[h]
    \centering
    \includegraphics[width=1\linewidth]{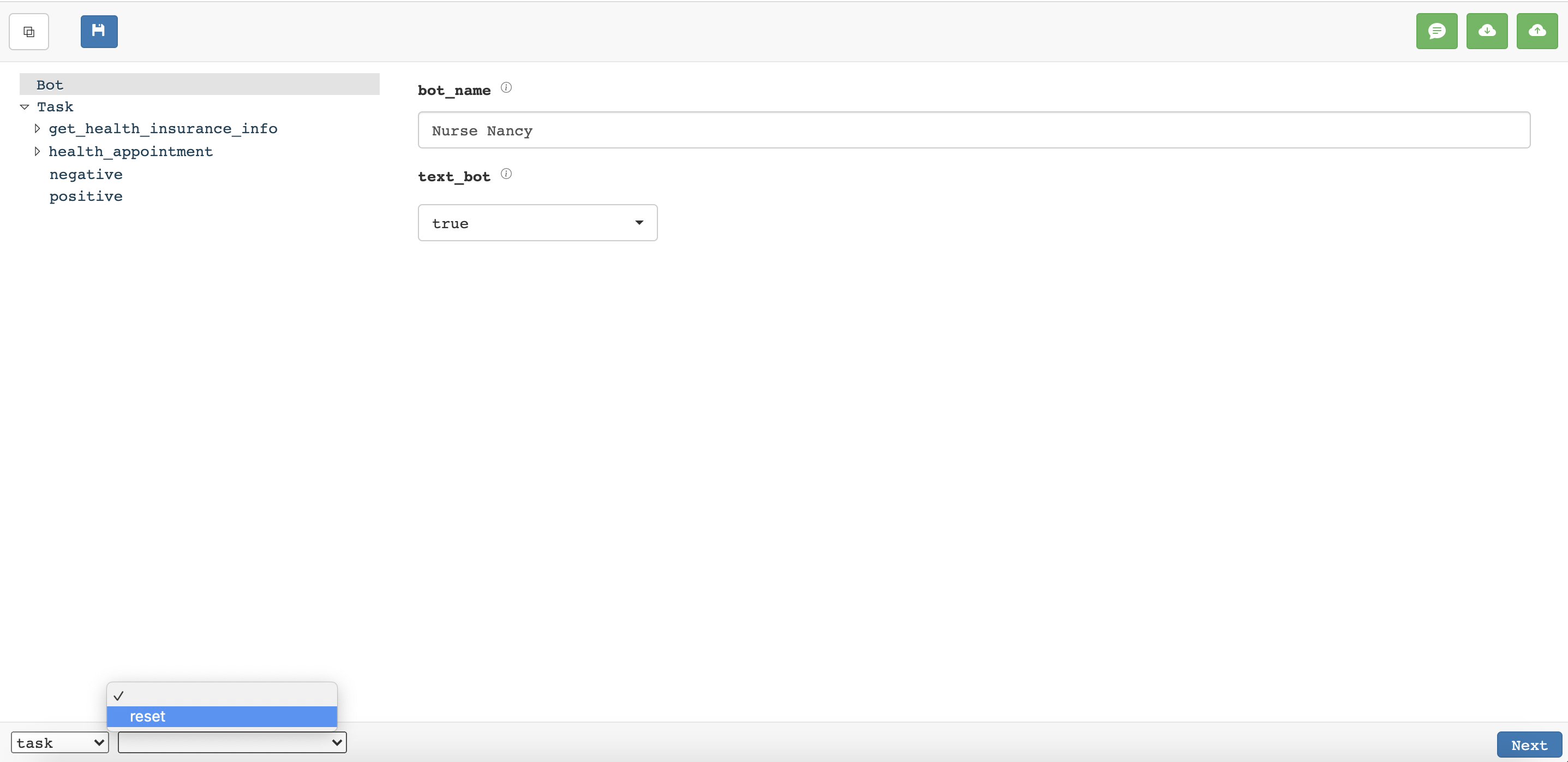}
    \caption{Converse config UI}
    \label{fig:configUI}
\end{figure}

We provide a config UI tool to create or edit the bot, see Figure \ref{fig:configUI}.

\begin{figure}[h]
    \centering
    \includegraphics[width=1\linewidth]{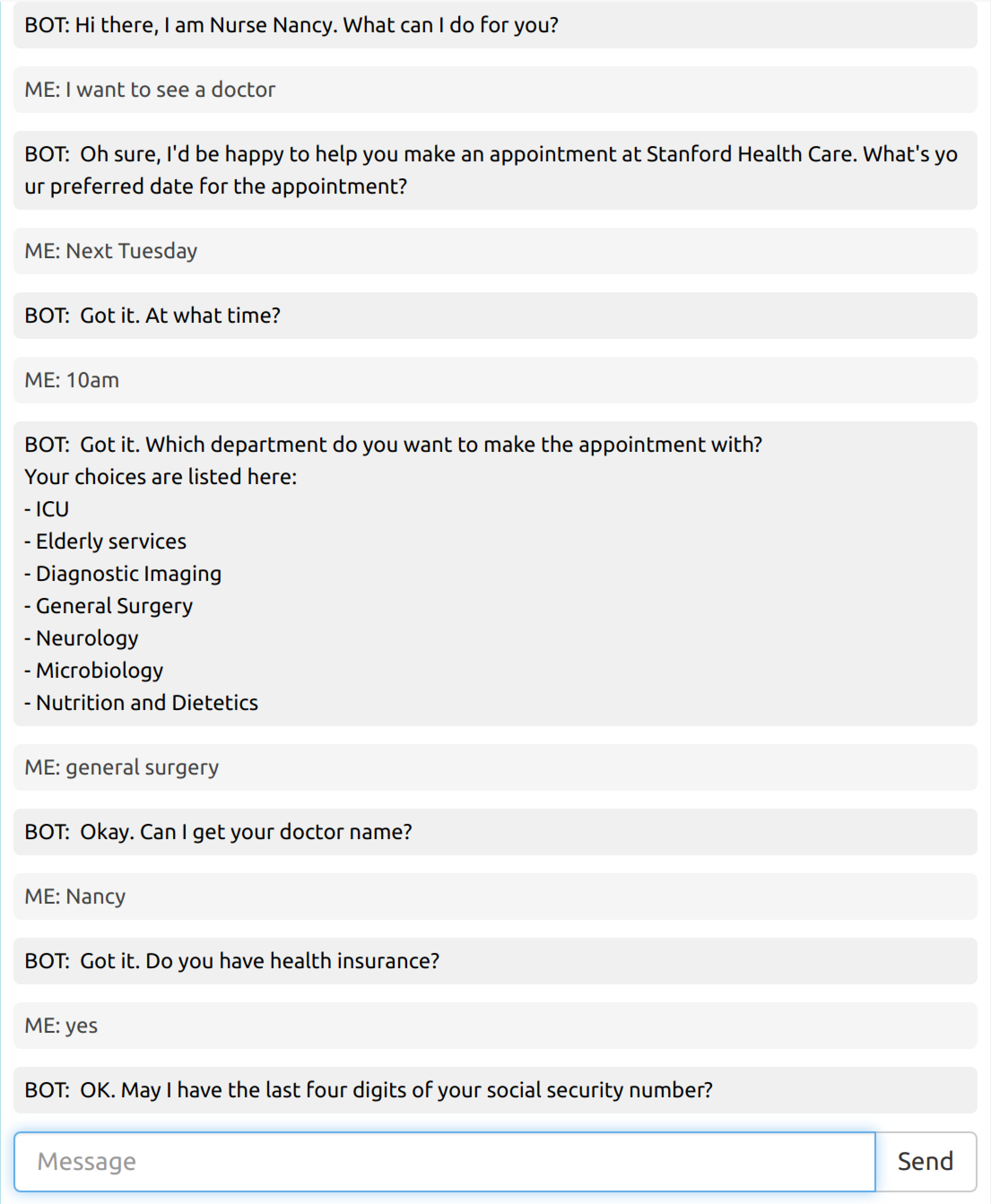}
    \caption{Example conversation of the health appointment bot}
    \label{fig:conversation}
\end{figure}

\begin{figure}[h]
    \centering
    \includegraphics[width=1\linewidth]{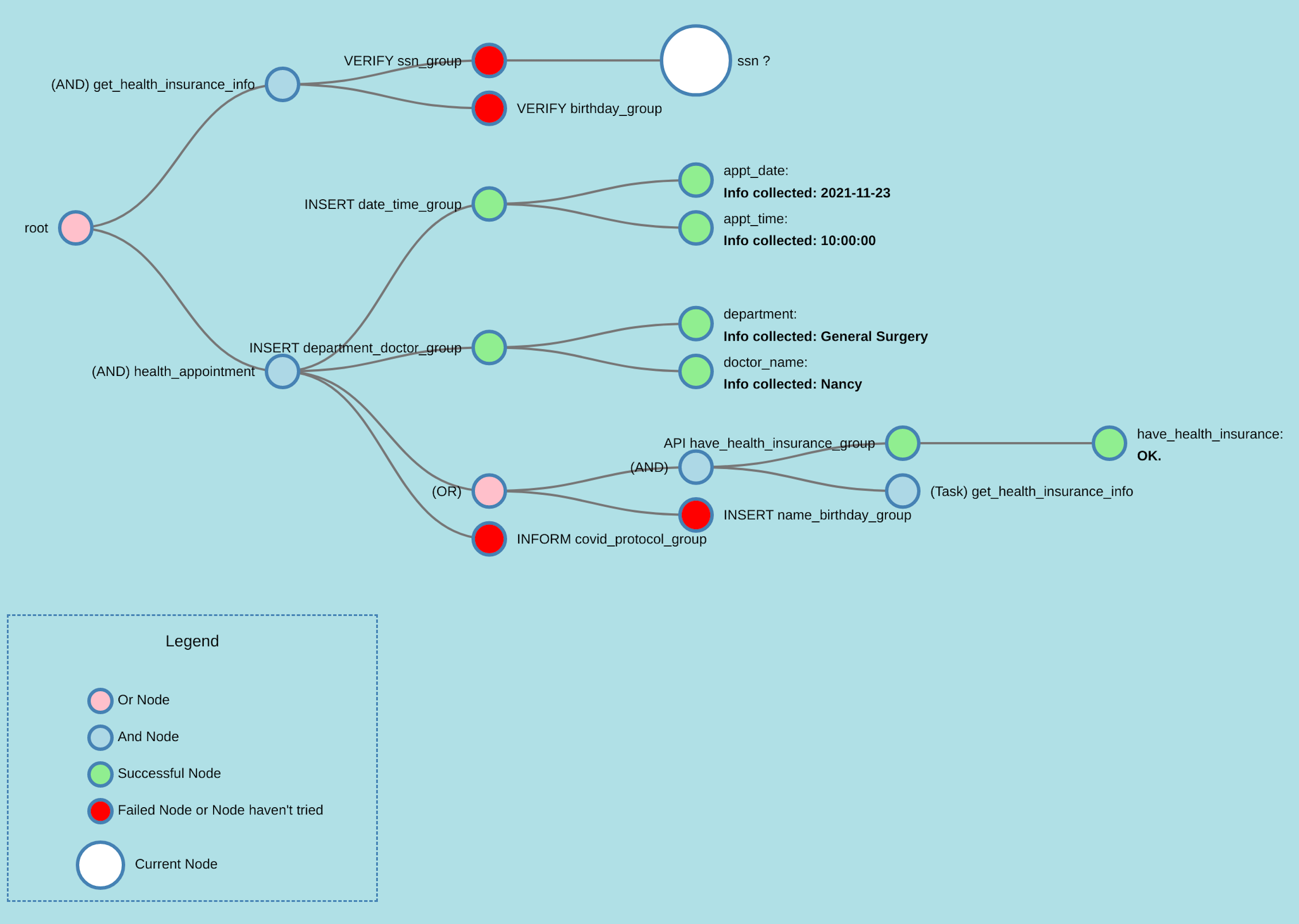}
    \caption{Task tree visualization of the health appointment bot}
    \label{fig:task_tree_vis}
\end{figure}

Here is an example conversation of the health appointment bot, see Figure \ref{fig:conversation} and \ref{fig:task_tree_vis}

\end{document}